\pdfoutput=1

\documentclass[12pt]{ouparticle}

\usepackage{natbib}
\usepackage{booktabs}
\usepackage[hidelinks]{hyperref}
\usepackage{multirow}
\usepackage[T1]{fontenc}
\usepackage[utf8]{inputenc}
\usepackage{microtype}
\usepackage{inconsolata}
\usepackage{amsmath}
\usepackage{amssymb}
\usepackage{comment}
\usepackage{xcolor}

\usepackage{lipsum}  
\usepackage{mwe}

\newcommand{\hcoal}{\texttt{H-COAL}}

\usepackage{array}
\newcolumntype{P}[1]{>{\centering\arraybackslash}p{#1}}

\begin{document}

\title{\hcoal{}: Human Correction of AI-Generated Labels for Biomedical Named Entity Recognition}

\author{%
\name{Xiaojing Duan, John P. Lalor}
\address{IT, Analytics, and Operations\\
University of Notre Dame}
\email{xduan@nd.edu, john.lalor@nd.edu}
}
\abstract{

With the rapid advancement of machine learning models for NLP tasks, collecting high-fidelity labels from AI models is a realistic possibility.
Firms now make AI available to customers via predictions as a service (PaaS).
This includes PaaS products for healthcare.
It is unclear whether these labels can be used for training a local model without expensive annotation checking by in-house experts.
In this work, we propose a new framework for Human Correction of AI-Generated Labels (\hcoal{}).
By ranking AI-generated outputs, one can selectively correct labels and approach gold standard performance (100\% human labeling) with significantly less human effort.
We show that correcting 5\% of labels can close the AI-human performance gap by up to 64\% relative improvement, and correcting 20\% of labels can close the performance gap by up to 86\% relative improvement. 
}

\date{\today}

\keywords{natural language processing; healthcare; healthcare analytics;\\ prediction uncertainty; named entity recognition; machine learning}

\maketitle
\section{Introduction}

%BioNLP workshop details: \url{https://aclweb.org/aclwiki/BioNLP_Workshop}

%Problem setting is that of an organization who relies on a 3rd party prediction as a service provider (PaaS) (or prediction on demand - POD) e.g., Amazon AI or Microsoft AI, etc. 
%We propose a method for Active Knowledge Distillation, so that local users can correct PaaS outputs to better inform distillation training.

AI models are becoming more accurate and are exceeding performance expectations across domains, including in areas such as law and medicine.
For example, ChatGPT recently passed the US Medical Licensing Exam \citep[USMLE, ][]{Kennedy_2023}.
Firms seek ways to incorporate these powerful AI models into their organizations to extract value.
However, these models require significant capital to obtain such impressive performance.
Recent advances have led to models that are so large and costly to train that they can only be trained and maintained by large organizations.
As a result, AI-as-a-service or prediction-as-a-service (PaaS) has become a new business offering for firms such as Google, Microsoft, and Amazon, among others~\citep{agrawal2018prediction}.
With PaaS, firms offer an application programming interface (API) to customers.
Firms can leverage PaaS output to build local applications.
For example, Kepro, a firm for healthcare management, developed an application where their review teams use Microsoft Text Analytics for Health to identify entities and correct them as needed \citep{microsoftazureKeproImprovesHealthcare2021}.
While this approach reduces development time and reduces data labeling costs, developers must still confirm the data that is being provided.

Customers can also use the API to generate predictions from black-box models based on their in-house data, making it possible to leverage state-of-the-art AI without expensive model training and storage.
For example, Microsoft's Azure Text Analytics for Health has a medical named entity recognition (NER) feature to extract medical entities from texts.
Microsoft notes that the platform is provided ``AS IS'' \citep{microsoftazureWhatTextAnalytics2023}.
Firms hoping to leverage the capabilities of such services still need to incorporate some degree of quality control when examining the provided outputs.

\begin{figure}
    \centering
    \includegraphics[width=\textwidth]{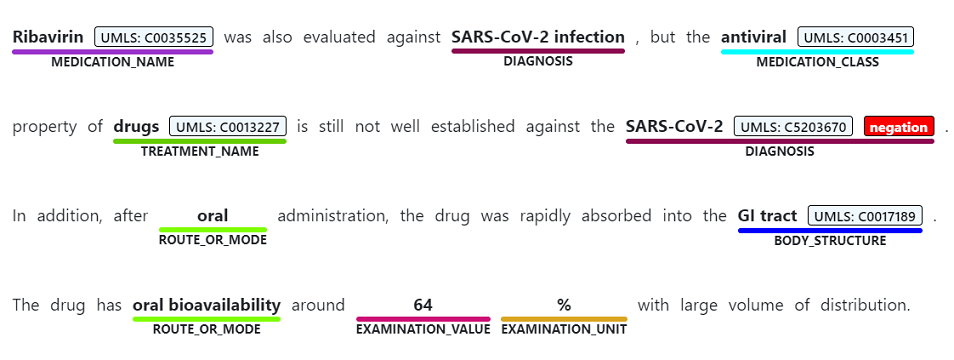}
    \caption[t]{An example of output from the Microsoft Azure Text Analytics for Health PaaS. For a given input text, the PaaS outputs identified entities \citep{microsoftazureWhatTextAnalytics2023}.}
    \label{fig:healthPaaS}
\end{figure}

While the proliferation of PaaS providers means more access to powerful AI predictions, it is unclear how well clients can trust the outputs of the API models~\citep{sushil2021we}.
For example, certain APIs return a prediction and a confidence score, but how such a confidence score is generated is not always transparent.
Therefore, if a firm wants to use the PaaS outputs (e.g., AI-generated labels) to train a local machine learning model, there needs to be a way to identify and correct errors made by the PaaS model in a fast and efficient manner.
Firms cannot manually check each instance; if they could, they would not require the services of the PaaS provider.
Therefore it is necessary to develop new methods for this emergent problem.
%Such methods will need to merge active learning and difficulty estimation, among other tasks.

This need is particularly salient in healthcare~\citep{chua2022tackling}.
Prediction uncertainty leads to the slow adoption of machine learning models in fields such as radiology.
Recent work has called for a better understanding of prediction uncertainty in healthcare machine learning applications and the development of new methods and metrics for error-tolerant machine learning.

One recently proposed framework for addressing noisy data is active label cleaning.
Active label cleaning is a process of correcting noisy labels~\citep{bernhardt2022active}.
It focuses on identifying and rectifying errors, inconsistencies, or noise in the labeled data, ensuring its accuracy and reliability, thereby improving the model's performance and generability. 
Active label cleaning can complement the existing noise-handling approaches by preserving the highly informative samples that otherwise could be disregarded and improving the quality of labeled data used for training and evaluation.
With data and labels being collected in a wide variety of ways, active label cleaning is a critical step to ensuring high-quality data that can be used to train machine learning models.

In this work, we present a framework for Human Correction of AI-Generated Labels (\hcoal{}).
Assuming a firm wants to obtain a large number of predictions via a PaaS provider, \hcoal{} can identify those generated labels that are most likely to be incorrect. 
These labels are routed to a human expert for review and correction if necessary.
Experiments on the i2b2 2010 named entity recognition dataset \citep[NER, ][]{uzuner20112010} indicate that by identifying and correcting as few as 1083 examples (5\% of the data), macro-F1 performance can improve by 2.3 absolute percentage points.
To reduce expensive expert annotation time, \hcoal{} can close the AI-expert labeling gap by 64\% relative improvement with as little as 5\% of the expected human annotation expenditure.

Our contributions in this work are: (i) A new framework, \hcoal{}, for identifying those AI-generated labels most likely to be wrong so that they can be corrected by humans, (ii) An empirical investigation of three possible ranking methods: LengthRank, EntityRank, and ConfidenceRank (iii) Empirical evidence that \hcoal{} improves local model training at significant cost savings.%\footnote{Code and data will be released upon publication.}

The rest of this work is structured as follows.
In Section \ref{sec:relatedwork} we discuss related work in the area.
In Section \ref{sec:hcoal}, we describe the \hcoal{} framework.
In Section \ref{sec:experiments}, we describe our experiments to validate \hcoal{}.
In Section \ref{sec:results}, we analyze our results.
In Section \ref{sec:conclusion} we discuss our results and limitations and look forward to future research opportunities in the area.

\section{Related Work}
\label{sec:relatedwork}

When firms, particularly in healthcare, are considering implementing machine learning models, uncertainty is a major issue.
There are a number of ways uncertainty can arise in a machine learning pipeline.
In this work, we leverage the classification of prediction uncertainty in healthcare presented by \cite{chua2022tackling}.
Specifically, the authors describe three types of uncertainty~\citep{chua2022tackling}:

\begin{enumerate}
\item
\textit{Out-of-distribution uncertainty} refers to uncertainty that is due to missing information in the training dataset. This could refer to data that is collected for one purpose and used to train a model for another related purpose, or data collection that does not have proper annotation guidelines, among other issues.
\item
\textit{Aleatoric uncertainty} refers to noisy training data. This could arise from poor annotations, ambiguity in the task and/or specific training examples, or issues with information extraction when creating the dataset to be labeled.
\item
\textit{Model uncertainty} refers to uncertainty that occurs during the model design and selection process. Hyperparameter selection, model or distribution configurations, etc. all affect the training and prediction process and introduce uncertainty.
\end{enumerate}

In this work, we are specifically interested in alleviating \textit{aleatoric uncertainty}.
What's more, this work introduces a new type of aleatoric uncertainty: label noise due to PaaS.
Although the framework of \cite{chua2022tackling} was introduced recently, since its introduction the leap that generative large language models (LLMs) such as ChatGPT have made in performance has made it possible to obtain high fidelity (but still noisy) labels directly from AI.
The framework already needs to be updated to account for this new source of potential uncertainty, which we address here.

In this work, we focus on PaaS for named entity recognition (NER). 
NER is an essential task in NLP that seeks to extract and classify named entities into predefined categories. 
The categories can be generic like Person, Organization, Location, Time, or tailored to a particular domain such as healthcare (e.g., Treatment, Test, Problem).  
By accurately recognizing and categorizing named entities, NER plays a crucial role in various NLP applications, such as information extraction, question-answering systems, text summarization, and more. 
It is a common type of PaaS, included in products such as Azure Text Analytics for Healthcare and Amazon Comprehend, among others.
NER has also been used in information systems as a critical component of design science artifacts~\citep{li2020theoryon,etudo2023ontology}.

A related stream of work is active learning~\citep{settles2008analysis} and human-in-the-loop learning~\citep{WU2022364}. 
%In particular, active learning for NER.
Prior work has shown that active learning is effective for both open domain NER~\citep{liu2022ltp,shen2017deep} and Bio-NER~\citep{chen2015study}. 
While active learning methods help identify candidate items for labeling by experts, here we seek to identify candidate items for error correction~\citep{rehbein2017detecting}, as all items have in fact been labeled (by the PaaS model).
%who proposed to use active learning for error detection in automatically generated labels. 
%However, their focus is different from ours. 
%While their goal is to improve the language data, our primary goal is to enhance the student model performance by identifying and correcting as few teacher-generated labels as possible.

%\citet{liu2022ltp} demonstrated that using an uncertainty-based active learning strategy could reduce the number of annotation tokens required by 20\% while maintaining competitive performance on both sentence-level accuracy and entity-level F1-score, across multiple Chinese NER tasks. 
%\citet{shen2017deep} incorporated incremental active learning during the training process of NER tasks and achieved nearly state-of-the-art performance using only 25\% of the original training data. 
%\citet{chen2015study} applied active learning to the clinical NER task and experimented with the uncertainty-based, diversity-based, and base-line sampling approaches. 

%They found the uncertainty-based strategy significantly saved annotation costs. 
%However, only a few studies have applied active learning for error detection/correction of AI-generated labels. 

\begin{figure}[h]
    \centering
        \subfigure[]{\includegraphics[width=0.45\textwidth]{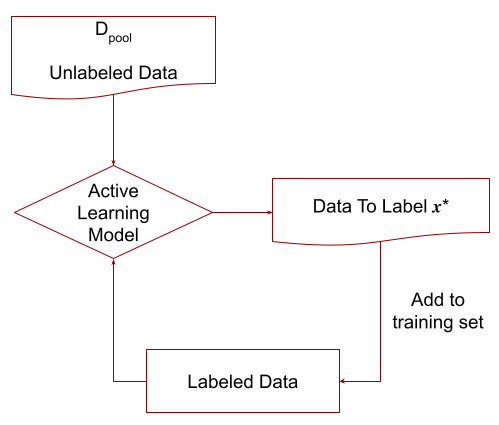}\label{fig:activelearning}}\hfill
        \subfigure[]{\includegraphics[width=0.45\textwidth]{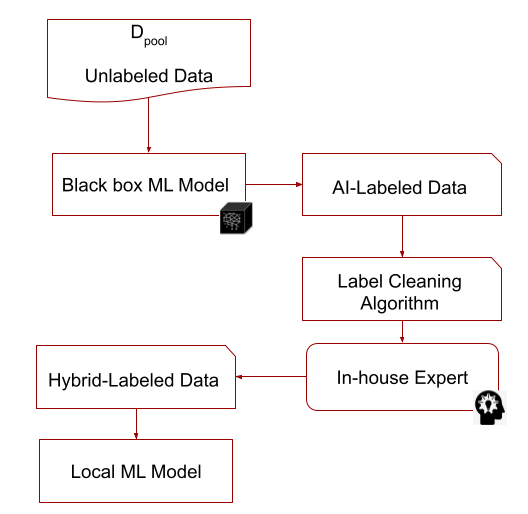}\label{fig:hcoal}}
    \caption{A comparison of active learning (\ref{fig:activelearning}) and our \hcoal{} framework (\ref{fig:hcoal}). The active learning process is iterative, where labeled data improves the local model and informs candidate data identification. In \hcoal{}, and in active label cleaning more broadly, all data points have (noisy) labels initially. The goal is then to identify and correct those incorrect labels.}
\end{figure}

Figure \ref{fig:activelearning} shows how active learning typically works. 
Typically, there is a large, \textit{unlabeled} pool of data.
An active learning module identifies those examples that should be labeled, and those examples are passed to a human oracle for labeling.
As this process continues, the training set grows, and performance improves, typically better than a random sampling procedure.
Critically, in our context, \textit{all of the data} is labeled initially.
However, we assume some percentage of that data is labeled incorrectly by the black-box PaaS service.
Therefore it is necessary to identify how many and which examples need to be checked (and possibly corrected) by the human expert.

Another related area is pre-labeling for medical annotation~\citep{lingren2014evaluating,gobbel2014assisted,wei2019cost}.
Pre-labeling assumes that the annotator will still annotate all examples, and the goal is time savings while retaining performance.
Here we only want the human to annotate a specific subset of the data, and keep the rest of the AI-generated labels.
What's more, circumstances such as fatigue and expertise limit the effectiveness of active-learning pre-labeling as humans work through a full dataset~\citep{wei2019cost}.

Prior work has investigated using GPT-3 for mixed human-AI annotation \citep{wang2021want}.
They note that performance was not appropriate for ``high-stakes'' cases such as medical data \cite[][p. 8]{wang2021want}.
Here we empirically demonstrate the gap between AI-generated labels and human-generated labels for Bio-NER and show that it can be efficiently reduced with targeted identification of data for relabeling.

Lastly, this work is related to the recent advances in reinforcement learning with human feedback (RLHF), which has been deployed to huge success with ChatGPT~\citep{christiano2017deep,daniels2022expertise}.
However, RLHF assumes that the model is available for fine-tuning based on human feedback.
In our setup, the model is a pure black box, and we cannot inject any additional supervision upstream.

\section{\hcoal{} Framework}
\label{sec:hcoal}
\subsection{Overview}

Figure \ref{fig:hcoal} shows an overview of the \hcoal{} framework. 
We assume there is a pretrained, black-box, PaaS model and a local, unlabeled candidate pool of data to be labeled.
The goal is to generate labels for this dataset such that a local machine learning model can be trained.
The PaaS model will label all unlabeled items, as inference is relatively inexpensive.
We consider this AI-labeled data the training set for our local model and the point at which we need to correct as many mistakes by the PaaS model as we can.
We first must determine which AI-generated labels are most likely to be wrong, so we can present them to a human expert (e.g., a clinician) for correction.
To do so we select a percentage of training examples from the data set most likely to contain errors.

%As a heuristic here, we could select the examples with the most labels, and assume (naively) that each label is equally likely to be wrong.
The firm's in-house expert then inspects the selected examples and corrects any incorrect labels.
Our final training dataset consists of (mostly) AI-generated labels and a sample of human-generated labels.
We then train the local model with the new labels.
We note here that this is not a knowledge distillation training procedure. 
We do not have ``gold standard'' labels with which to perform distillation. 
Instead, we have the AI-generated labels (which we consider ground truth) and the subset of those labels that have been corrected (where necessary) by the human expert.

\subsection{Identifying Possible Errors}

We propose three methods for identifying examples for error correction.
The first is example length (\textbf{LengthRank}).
Prior work typically uses example length as a proxy for difficulty \citep[e.g., ][]{platanios2019competence, lalor2020dynamic}.
We assume that the longest examples will be the ones with the most errors and therefore the ones that we should route to the human expert for correction.
For each example $x_i$ we can calculate the length (number of words) $l_i = \text{len}(x_i)$.
LengthRank ($r_L$) is then the ordered set of examples from longest to shortest:

\begin{equation}
%    l_i &= \text{len}(x_i) \\
    r_L = \{x_1, \dots , x_N \vert l_1 > l_i > l_N\}
\end{equation}

Our second heuristic is the number of entities in an example (\textbf{EntityRank}).
We assume that if an example has more entities to be labeled, then it is more likely that one or more of those entities will have errors.
For each example $x_i$ we count the number of entities identified by the PaaS model ($e_i = \sum_i^{l_i} \mathbb{I} [ \text{ent}(t_i) ]$), where $\mathbb{I}[x]$ is the \textit{indicator function} that returns $1$ if $x$ is true and $0$ if $x$ is false.
EntityRank ($r_E$) is the ordered set of examples from those with the most entities to those with the fewest entities:

\begin{equation}
%    e_i &= \sum_t^{x_i} \mathbb{I} [ \text{ent}(x_i) ]\\
    r_E = \{x_1, \dots , x_N \vert e_1 > e_i > e_N\}
\end{equation}

Our third ranking relies on the black box model's confidence outputs (\textbf{ConfidenceRank}).
Typically, each PaaS model will output a confidence score with its predictions.
While the nature of how these scores are calculated is not always known, we can use them as intended, to indicate the model confidence in a given label.
Therefore, our final ranking looks at the least confident labels first.
For each example $x_i$ we identify the entity with the lowest confidence score from the PaaS model ($c_i = \min (\text{conf}(t_i))~~\forall t_i \in x_i \text{ where } \text{ent}(t_i) \text{ is True}$).
ConfidenceRank ($r_C$) is the ordered set of examples from those with the lowest confidence result to those with the highest confidence result:

\begin{equation}
    %c_i &= \min \text{conf}(t) \forall t \in x_i\\
    r_C = \{x_1, \dots , x_N \vert c_1 < c_i < c_N\}
\end{equation}

\section{Experiments}
\label{sec:experiments}

To test the efficacy of \hcoal{} we conducted an experiment on biomedical named entity recognition (Bio-NER).
Our goal was to replicate the environment of a hospital or other medical practice that has partnered with a large PaaS provider to obtain labels for their (unlabeled) pool of data. 
The experiment was conducted using the i2b2 2010 dataset~\citep{uzuner20112010}, consisting of 37,105 examples with gold-standard labels in the beginning-inside-outside (BIO) format. 
We fine-tuned the Clinical-BioBERT ~\citep{alsentzer2019publicly} model and used the result as the black box PaaS model in our experiment. 
The Clinical-BioBERT model was initialized from BioBert and has shown improved performance on Bio-NER tasks compared to its predecessor.

To start, we fine-tuned ClinicalBioBERT on a small sample of i2b2 data in order to prime the model for the task of Bio-NER with our label set.
Note that this step would not typically occur in the real-world scenario we are attempting to emulate.
We assume that the PaaS provider has a mechanism in place for ensuring that the black-box model can handle the task at hand. 
We simulate that here with a brief fine-tuning of an off-the-shelf model.
We fine-tuned Clinical-BioBERT with a batch size of 16, a maximum sequence length of 150, a learning rate of 2e-05, and two epochs. 
We used the fine-tuned model to generate labels for a larger portion of the dataset.
Table \ref{tab:dataset} shows how many i2b2 examples were used for each stage of this process.

\begin{table}[h]
\small
\centering
  \begin{tabular}{lc}
    \toprule
     {Task} & Example Count\\
    \midrule
    Fine-tune Clinical-BioBERT ($D_{PaaS}$) & 5413 \\
    Generate labels for fine-tuning ($D_{pool}$)& 21651\\
    Local Test Set ($D_{test}$) & 10041\\
    \bottomrule
  \end{tabular}
  \caption{A description of how the data was split for our experiments. We use a relatively small number of examples to fine-tune our (simulated) PaaS model ($D_{PaaS}$). Our local dataset to be labeled ($D_{pool}$) is large enough to ensure relatively good performance. We apply \hcoal{} to $D_{pool}$, and use the local test set ($D_{test}$) for evaluation.}
  \label{tab:dataset}
\end{table}

The PaaS-labeled data is the input for \hcoal{}.
For each of our ranking schemes, we sampled the top 5\%, 10\%, and 20\% of the data based on each ranking strategy for label correction. 
For these examples, we simulated review by an expert and used the true labels as obtained from the original i2b2 dataset.
We then fine-tuned a \textit{new} Clinical-BioBERT model with our local, mixed-label data.
We used this new Clinical-BioBERT model to emulate a local model downloaded from a model repository such as HuggingFace \citep{wolf-etal-2020-transformers}.

As our baselines, we include a fine-tuning process where only the AI-generated labels are used as a performance floor, a strategy that randomly samples examples for review, and a fine-tuning with the full human-labeled dataset as a performance ceiling.
We hope to approach the performance of fully human-labeled data with a small percentage of error correction for efficiency.
For each fine-tuned model, we report F1 scores for each entity type, as well as micro average F1, macro average F1, and weighted average F1 scores.
Micro average F1 calculates a global F1 by looking at the overall data.
Macro F1 calculates scores for each label and then takes their mean.
Weighted F1 is similar to macro averaged, but takes a weighted mean based on the number of examples for each class.

\begin{comment}
\begin{table}[t]
\centering
  \begin{tabular}{lc}
    \toprule
     {Entity Type} & Entity Count\\
    \midrule
    Problem & 8461 \\
    Test & 4962\\
    Treatment & 8029\\
    \cmidrule{2-2}
    Total & 21452\\
    \bottomrule
  \end{tabular}
  \caption{Type and count of entities identified during the test.}
  \label{tab:entity}
\end{table}
\end{comment}

\section{Results}
\label{sec:results}
We first report the results of the quantitative comparisons between different sampling strategies and then describe a qualitative analysis of the ConfidenceRank outputs.

\subsection{Quantitative Comparisons}
Table \ref{tab:Results} shows the results of different sampling strategies. 
We first note that while the gap between AI-generated labels and gold-standard labels is relatively small, it does exist. 
The gap varies across entity types, with Problem being the largest (6\%) and Test being the smallest (2.9\%). 
This shows that there is an AI-human gap in the labels, leaving room for improvement by label correction from an in-house expert.

For the Length and Entity ranking strategies, correcting 5\% of the labeled data does not affect the overall performance of the local models. 
Even when correcting the top-20\% of examples, improvement is relatively small.
%However, when increasing the sample size to 20\%, we observe performance improvement. 
At 20\% correction, the Length ranking strategy improved the macro average F1 score by 2.2\% and the Entity ranking strategy improved it by 1.9\%.

The Confidence ranking strategy exhibits an improvement of 2.7\% after correcting only 5\% of the sample. 
As one might expect, correcting those examples where the AI model is least confident leads to the best performance. As the number of corrected samples increases, the performance continues to improve. By correcting only 20\% of the data, macro average F1 scores approach the ceiling of annotating all of the data. 
Closing the gap between \hcoal{} and fully-labeled data to 0.5\% for macro-F1.

\begin{table}[h!]
\small
\centering
\begin{tabular}{lccccP{1cm}P{1cm}P{1.5cm}}
\toprule
Budget & Ranking & Problem&Test&Treatment&Micro Avg. &Macro Avg.& Weighted Avg.\\
&&($n=8461$)&($n=4962$)&($n=8029$)&\multicolumn{3}{c}{($n=21,452$) }\\
\midrule
\multicolumn{2}{l}{0\% (AI Labels) } & 83.8&84.8&85.7&84.7&84.8&84.7\\
\cmidrule{3-8}
\multirow{4}{*}{5\%}&Random & 83.8 & 85.2 & 85.8  & 84.9 & 84.9  & 84.9 \\
&$r_L$  & 83.9 & 84.9 & 85.2  & 84.6 & 84.6  & 84.6 \\
&$r_E$  & 83.5 & 85.6 & 87.3 & 85.4 & 85.5 & 85.4 \\
&$r_C$ &\textbf{86.9} & \textbf{86.4} & \textbf{88.1} & \textbf{87.2} & \textbf{87.1} & \textbf{87.2} \\
\cmidrule{3-8}
\multirow{4}{*}{10\%}& Random& 84.2 & 85.2 & 87.5 & 85.6 & 85.6 & 85.6 \\
&$r_L$ & 85.0 & 86.1 & 87.0 & 86.0 & 86.0 & 86.0 \\
&$r_E$ &84.6 & 85.0 & 87.3 & 86.0 & 85.6 & 85.7 \\
&$r_C$ & \textbf{87.2}& \textbf{86.9} & \textbf{88.4} & \textbf{87.6} & \textbf{87.5} & \textbf{87.6} \\
\cmidrule{3-8}
\multirow{4}{*}{20\%}& Random& 85.2 & 86.8 & 87.4 & 86.4 & 86.5 & 86.4 \\
&$r_L$ & 86.0 & 85.6 & 88.6 & 86.9 & 86.7 & 86.9 \\
&$r_E$  &85.6 & 85.8 & 87.9 & 86.5 & 86.4 & 86.5 \\
&$r_C$   &\textbf{87.4} & \textbf{87.1} & \textbf{89.0} & \textbf{88.0} & \textbf{87.9} & \textbf{88.0} \\
\cmidrule{3-8}
\multicolumn{2}{l}{100\% (Gold Labels) } & 88.8 & 87.3 & 89.2 & 88.6 & 88.4 & 88.6 \\
\bottomrule
\end{tabular}
\caption{Entity level and average F1 scores of the local model trained with different configurations of mixed AI and gold labels. Percent values for each ranking strategy indicate that the top-ranked $N$\% of AI-generated labels were checked and, if necessary, replaced. Best performing mixed strategy at each budget level is \textbf{bolded}.
}
\label{tab:Results}
\end{table}

\begin{figure}
    \subfigure[Number of entities requiring correction selected by different sampling strategies.]{\includegraphics[width=0.45\textwidth]{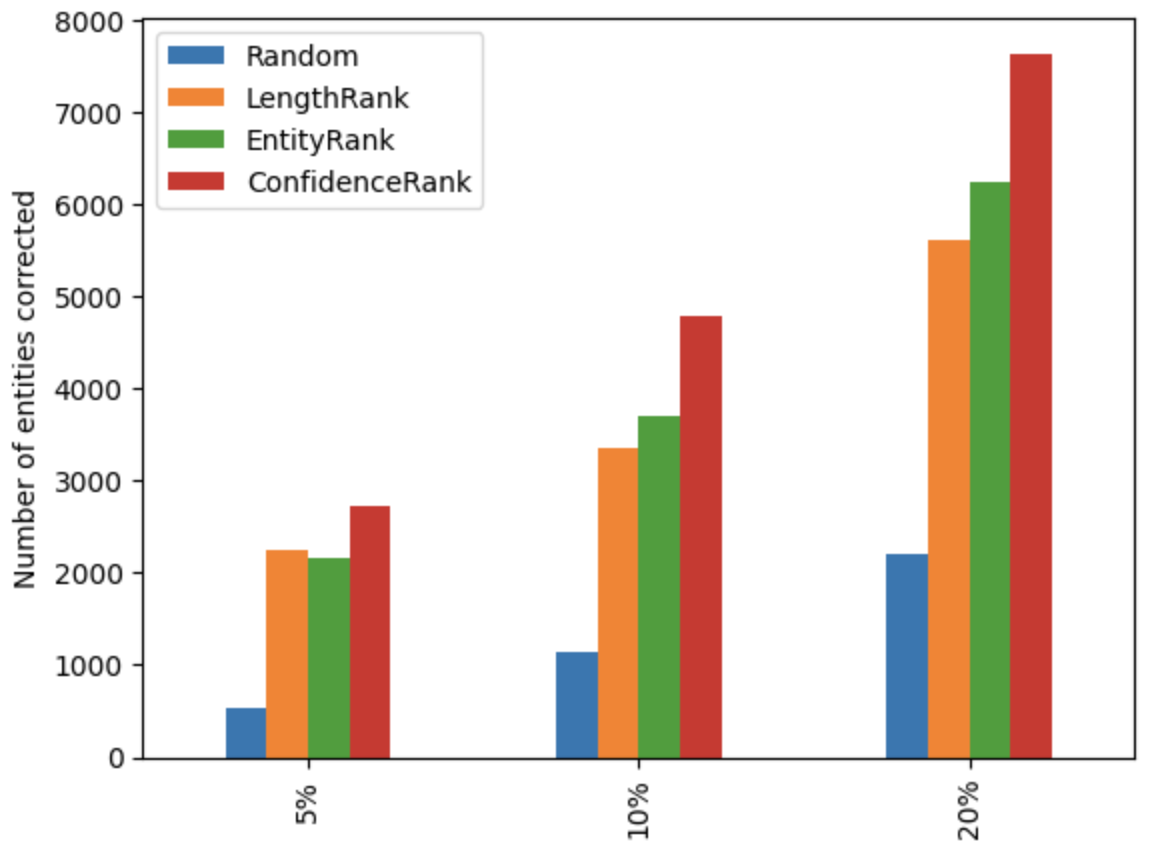}\label{fig:entities_corrected}}
    \hfill
    \subfigure[Number of examples requiring correction selected by different sampling strategies.]{\includegraphics[width=0.45\textwidth]{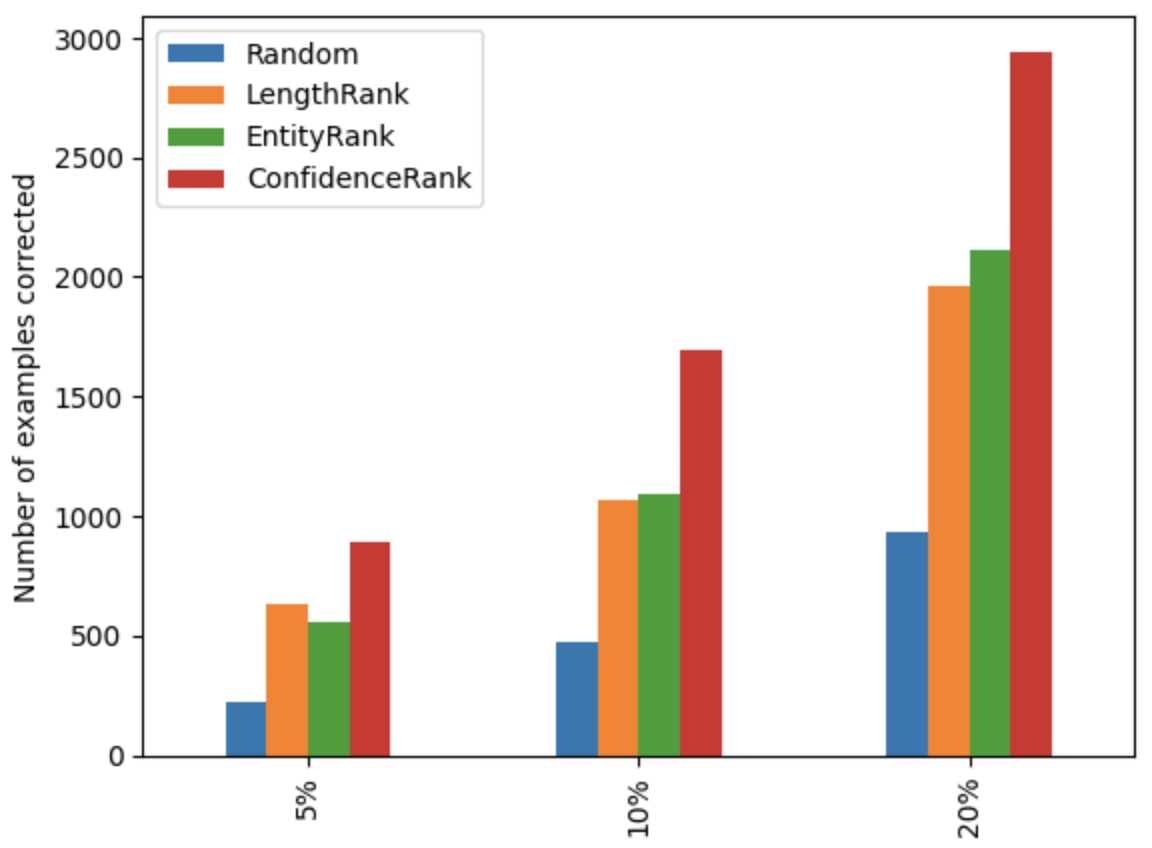}\label{fig:sentences_corrected}}
\end{figure}

We also compared the number of entities and examples that require correction in the samples selected by different sampling strategies. 
As shown in Figure \ref{fig:entities_corrected}, ConfidenceRank identified the most entities that require correction. 
Similarly, Figure \ref{fig:sentences_corrected} shows ConfidenceRank also identified the most examples requiring correction. 
These findings suggest that ConfidenceRank is the most effective sampling strategy in terms of optimizing the utilization of human experts' time by focusing on examples or entities that need correction. 

%\caption{Comparison of the number of entities (\ref{fig:entities_corrected}) and examples (\ref{fig:sentences_corrected}) requiring correction identified by different ranking strategies. Percent values indicate the top-ranked $N$\% of AI-generated labels checked by different ranking strategies. }
%\label{fig:graphs}
%\end{figure}

\begin{table}[h!]
\small
\centering
\begin{tabular}{lccccP{1cm}P{1cm}P{1.5cm}}
\toprule
Budget & Ranking & Entities Corrected & Entities Identified & Percentage Corrected \\
\midrule
\multirow{4}{*}{5\%}&Random & 531 & 3,748 & 14.2 \\
&$r_L$  & 2,245 & 24,879 & 9.0  \\
&$r_E$  & 2,166 & 19,814 & 10.9 \\
&$r_C$ & 2,729 & 18,290 & 14.9  \\
\midrule
\multirow{4}{*}{10\%}& Random& 1140 & 7,979 & 14.3 \\
&$r_L$ & 3,366 & 35,942 & 9.4  \\
&$r_E$ & 3,708 & 32,541 & 11.4  \\
&$r_C$ & 4,781 & 33,641 & 14.2  \\
\midrule
\multirow{4}{*}{20\%}& Random& 2,209 & 15,558 & 14.2 \\
&$r_L$ & 5,603 & 53,740 & 10.4 \\
&$r_E$  & 6,249 & 51,224 & 12.2 \\
&$r_C$  & 7,643 & 56,439 & 13.5 \\
\bottomrule
\end{tabular}
\caption{Percentage of entities requiring correction selected by different sampling strategies.
}
\label{tab:perc_entities_corrected}
\end{table}

\begin{figure}
    \centering
    \includegraphics[width=0.45\textwidth]{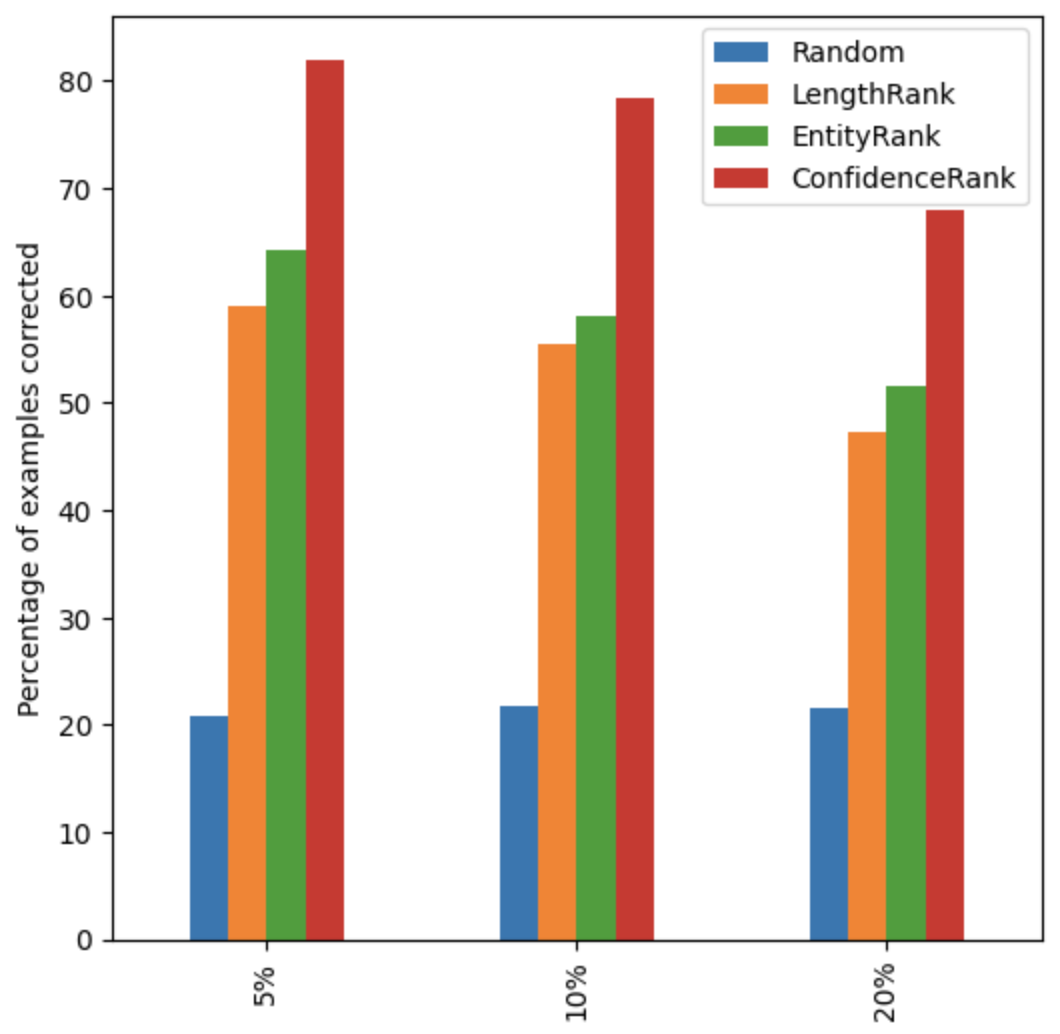}
    \caption{Percentage of examples requiring correction selected by different sampling strategies.}
    \label{fig:perc_sentences_corrected}
\end{figure}

Additionally, we conducted a comparison of the percentage of entities and examples requiring correction among different sampling strategies. Table \ref{tab:perc_entities_corrected} presents the percentage of entities requiring correction in different sampling strategies. As it shows, ConfidenceRank selected the highest percentage of entities requiring correction when the sampling size is 5\%. When the sampling size is 10\% and 20\%, Random sampling selected a higher percentage of entities requiring correction. However, it is worth noting that the total counts of selected entities for Random sampling are significantly lower (7,979 and 15,558) compared to the other strategies with the same sampling size. This suggests that the higher percentage of entities selected by Random sampling comes at the cost of fewer entities being chosen overall. When analyzing the percentage of examples requiring correction in the samples selected by different strategies, it is evident from Figure \ref{fig:perc_sentences_corrected} that ConfidenceRank consistently selects the highest percentage per sample size. This finding highlights the effectiveness of the ConfidenceRank strategy in identifying examples that need correction compared to other sampling strategies.

\subsection{Qualitative Analysis}
We next present a qualitative analysis of ConfidenceRank outputs (Table \ref{tab:Qualitative_Results}). 
%We selected examples from the top 5\%, 15\%, and 20\%  ranked sentences. 
The first row shows the highest-ranked example, where the black box model incorrectly identified \textbf{\emph{hydorxyurea}} as I-PROBLEM with a confidence score of 0.196. %, whereas the gold label is I-TREATMENT. 
%In the second row, we see a sample ranked within the top 5\%, where the black box model labeled \textbf{\emph{noxious}} as B-TREATMENT with a confidence score of 0.255, while its gold label is B-TEST. 
The second row shows another example ranked within the top 5\%. 
In this example, the black box model identified \textbf{\emph{A}} and \textbf{\emph{relook}} as B-TEST with a confidence score of 0.31 and  I-TEST with a confidence score of 0.61, respectively. 
However, the i2b2 gold standard doesn't have labels applied to those two tokens. 
These examples highlight that the confidence scores reflect the black box model's level of uncertainty about its output accuracy. 
Correcting examples such as these lead to significant improvements in the local model's performance. 
%These results also indicate that we can achieve near-ceiling performance by correcting only 5\% of the mislabeled samples identified by our ConfidenceRank strategy. 

The third row of Table \ref{tab:Qualitative_Results} illustrates an example ranked in the top 10\% using ConfidenceRank. 
Here, the black box model correctly labeled \textbf{\emph{scarring}} as B-PROBLEM with a confidence score of 0.534. 
This example demonstrates that even when the black box model is not very confident about its output, it can still be correct. 
%This finding may explain why increasing the label correction sample from 5\% to 10\% didn’t improve the local model's performance significantly. 
The last row shows a sample ranked in the top 20\%, where the black box model labeled \textbf{\emph{inadequate}} as B-PROBLEM with a high confidence score of 0.71, while its gold label is O. 
Although \textbf{\emph{inadequate}} can indicate a deficiency, which could be considered a problem, the gold label did not indicate a problem. 
This shows that confidence is not enough to identify all possible issues; future work on better ranking methods could further improve performance.
%This highlights the complexity of the task and the subjective nature of labeling, which can make it challenging for an AI model. 
%This can also explain why increasing the label correction sample to 20\% did not lead to significant improvement in the local model's performance. 

\begin{table*}[th]
\small
\centering
\begin{tabular}{ll}
\toprule
Example & \\%Gold Label & AI Label (Confidence) & ConfidenceRank \\
\midrule
\multirow{4}{7.5cm}{He had been noting night sweats, increasing fatigue, anorexia, and dyspnea, which were not particularly improved by increased transfusions or alterations of \textbf{\emph{hydorxyurea}}.} & Gold Label: I-TREATMENT\\
&  AI Label (Confidence): I-PROBLEM (0.196) \\
& $r_C$: 1 \\
&\\
%\midrule
%No withdrawal to \textbf{\emph{noxious}} stimuli. & B-TEST & B-TREATMENT (0.255) & 3\\
\midrule
\multirow{4}{7.5cm}{\textbf{\emph{A relook}} several days later led to a repeat percutaneous transluminal coronary angioplasty.} & Gold Label: O, O \\
&  AI Label (Confidence): B-TEST (0.31)\\
& \hspace{3.65cm} I-TEST (0.61) \\
& $r_C$: 32 \\
\midrule
\multirow{4}{7.5cm}{Chest x-ray revealed moderate cardiomegaly with no clear interstitial or alveolar pulmonary edema and chronic atelectasis and/or \textbf{\emph{scarring}} at both lung bases.} & Gold Label: B-PROBLEM \\
& AI Label (Confidence): B-PROBLEM (0.534) \\
& $r_C$: 2130 \\
&\\
\midrule
\multirow{4}{7.5cm}{However, the study was limited due to \textbf{\emph{inadequate}} PO contrast intake by the patient; it did show a question of a cecal cystic lesion verse normal loop of bowel.} & Gold Label: O  \\
& AI Label (Confidence): B-PROBLEM (0.71) \\
& $r_C$: 4245 \\
&\\
\bottomrule
\end{tabular}
\caption{Samples showing the gap between AI-generated labels and gold-standard labels.}
\label{tab:Qualitative_Results}
\end{table*}

\section{Conclusion}
\label{sec:conclusion}

\subsection{Summary of Results}
With the proliferation of AI and its improved performance, there are more opportunities than ever to leverage predictive modeling and generative tools such as ChatGPT~\citep{christiano2017deep,daniels2022expertise}.
However, especially in healthcare, care is needed to ensure that the outputs are valid and can be used by local experts.
In this work, we present \hcoal{}, a framework for human correction of AI-generated labels from PaaS providers.
By identifying examples most likely to be incorrect and routing them to an expert for correction, we can reduce the gap between AI-generated labels and fully-human labels by up to 64\% when only correcting 5\% of examples.
%we can see improvements of up to \todo{N 4} \% on various NER benchmark datasets. 

\subsection{Implications for Research and Practice}
This work has several implications for research and practice.
For research, we present a new framework for active label cleaning that relies on labels obtained from a black-box PaaS model.
This setup typically does not allow for explainable AI approaches, as the PaaS model is owned by a separate entity.
Typically the only information available besides the prediction is a confidence score.
This research can lead to a new stream focusing on the best way to design metrics for scoring PaaS outputs so that in-house experts know what to label.
In addition, new work in optimization can investigate the trade-offs between large-scale, potentially imperfect labeling via PaaS and more bespoke, local, expert-drive human annotations for different machine learning tasks based on budget, degree of difficulty, and other factors.

In terms of practical implications, firms can use \hcoal{} as part of their PaaS strategy.
\hcoal{} can alleviate concerns around errors in PaaS output by identifying those labels that need to be reviewed by in-house experts.
As our results show, not all examples that are identified for review are corrected, but when using ranking systems such as ConfidenceRank, firms can identify and correct more errors than random sampling and other more naive procedures.

\subsection{Limitations and Future Work}

This work has several limitations that make interesting avenues for future work.
First, the ConfidenceRank approach relies on the external confidence value from the black box PaaS model.
How this value is calculated is typically unknown.
Future work should investigate other ranking metrics that can be used with \hcoal{} to further improve the framework's predictive performance.
For example, a ranking approach that scores the input in terms of language perplexity or other PaaS-agnostic metrics would allow for flexible application of \hcoal{} that does not rely on the PaaS provider's confidence ranking.
%ConfidenceRank assumes that the PaaS model will output a reliable confidence score.
For example, with a language model, we can estimate the probability of a given example given that language model.
Less likely examples may have more errors than others.
Another method is readability scores such as Flesh-Kincaid.
It may be the case that more readable examples have more errors if they are too simple and missing medical context.
However, the opposite could also be true.

Second, this work simulates the described scenario of a hospital or medical firm leveraging PaaS systems. 
We simulate the scenario in order to carefully assess the framework, but future work should incorporate actual experts to better understand how and where PaaS systems are making errors. 
A more detailed, comprehensive assessment of how and why PaaS models fail for certain types of healthcare data would be beneficial for both practitioners using these systems and also for researchers seeking to understand and improve these models.

Another dimension of this work to consider is the ethical dimension.
The outputs of any BioNLP task should be carefully inspected by a medical expert before being used for medical decision-making.
In this work, we aim to improve the labeling process but do not endorse using any output from these models in a medical decision-making context without expert inspection.
As PaaS offerings become more cost-effective and performant for healthcare, more medical professionals and developers will come to rely on them for generating predictions for internal data.
It is necessary to have a framework in place for identifying and correcting errors from PaaS models to alleviate aleatoric uncertainty ~\citep{chua2022tackling}.

%\section*{Acknowledgements}

%\singlespacing
\bibliography{anthology,custom}
\bibliographystyle{acl_natbib}

%\appendix

%\section{Example Appendix}
%\label{sec:appendix}

%This is a section in the appendix.

\end{document}